\documentclass[twocolumn]{article}

\usepackage{amsmath}
\usepackage{graphicx}
\usepackage[utf8]{inputenc}
\usepackage{url}
\usepackage{booktabs}
\usepackage{multirow}
\usepackage{tikz}
\usepackage{subcaption}
\usepackage[colorlinks=true,linkcolor=blue,citecolor=blue,urlcolor=blue]{hyperref}

\usetikzlibrary{positioning, arrows.meta, calc}

\title{Frozen Vision Transformers for Dense Prediction\\on Small Datasets: A Case Study in Arrow Localization}

\author{Maxwell Shepherd\\\texttt{msheph15@jh.edu}\\{\small\url{https://github.com/maxwellshepherd/frozen_vision_transformers_for_archery_scoring}}} 

\date{}

\begin{document}

	\maketitle

	\section{Introduction}

	\paragraph{Background and Problem Statement}
	In competitive archery governed by World Archery regulations \cite{worldarchery}, archers shoot arrows at standardized target faces and record scores based on the concentric ring in which each arrow lands. The standard 40\,cm indoor target face consists of ten scoring rings, each 20\,mm wide, with point values ranging from 10 (innermost) to 1 (outermost). Scoring follows the line-cutting rule: if the arrow shaft touches a ring boundary, the higher of the two adjacent scores is awarded.

	Manual scoring records only a scalar point value per arrow, discarding the spatial information embedded in the physical target face. The two-dimensional distribution of arrow impacts contains information valuable for athlete development that conventional scoring does not capture: grouping tightness quantifies an archer's precision, the centroid of the group reveals systematic aiming bias, and tracking these spatial statistics over time can identify performance trends. Extracting this information manually from a target photograph is tedious and impractical at scale.

	We propose a deep learning system for automated detection and scoring of arrow punctures on 40\,cm indoor archery target photographs. Given a photograph of such a target, the system (1) detects every arrow puncture, (2) localizes each arrow shaft center with sub-millimeter accuracy in a canonical coordinate system relative to the target center, and (3) computes per-arrow scores under official line-cutting rules. Because the system recovers precise arrow positions rather than only point values, it enables downstream computation of grouping statistics, directional bias, and other spatial metrics useful for skill development. The present evaluation is limited to a single target type from a single venue; extension to other target faces or shooting environments would require additional validation.

	\begin{figure}[t]
		\centering
		\includegraphics[width=\columnwidth]{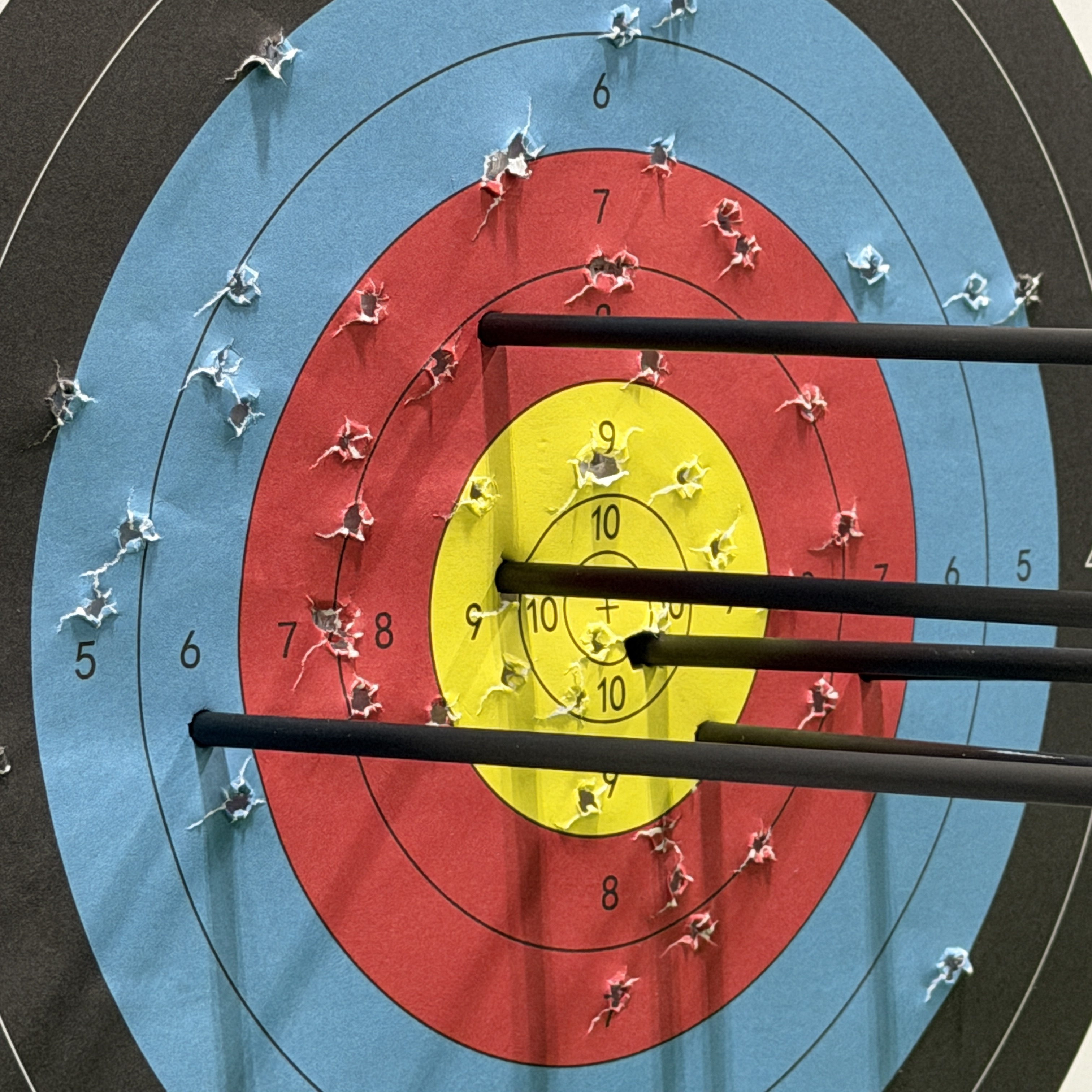}
		\caption{A 40\,cm indoor archery target face with arrow punctures across the scoring rings.}
		\label{fig:target}
	\end{figure}

	This task presents several challenges. First, the training dataset is small, on the order of dozens of images rather than the thousands more typical of supervised deep learning. Second, input photographs can contain perspective distortion from imperfect camera angles, causing circular scoring rings to appear as ellipses. Third, arrow punctures are small relative to the target face, creating an extreme class imbalance between foreground and background pixels. Fourth, compound punctures, where multiple arrows strike nearly the same location, produce overlapping tears that are difficult to resolve individually.

	To address these challenges, we combine a classical computer vision rectification pipeline with a modern transfer learning architecture. A color-based geometric rectification stage removes perspective distortion and maps all images into a canonical coordinate system before any learning occurs. A frozen self-supervised vision transformer (DINOv3 \cite{dinov3}) provides rich visual features without overfitting on a small dataset. Guided feature upsampling via AnyUp \cite{anyup} bridges the resolution gap between transformer patches and the sub-millimeter localization our task requires. Finally, lightweight CenterNet-style \cite{centernet} detection heads predict arrow center heatmaps and sub-pixel offsets, with only a few million trainable parameters out of hundreds of millions total.

	\paragraph{Related Work}
	Prior work on automated archery scoring is limited. Kim et al.\ \cite{kim2025} proposed an image-based system for outdoor archery that combines classical computer vision for target normalization with a convolutional network for heatmap-based arrow localization, achieving a mean error of 1.77\,mm. Their approach relies on a fully supervised convolutional network trained from scratch, requiring more data than our setting permits. We instead leverage self-supervised foundation model features, employ guided upsampling for high-resolution predictions, and introduce a specific rectification pipeline tailored to indoor target faces.

	CenterNet \cite{centernet} represents objects as single center points in a heatmap, trained with a penalty-reduced focal loss that downweights easy-negative predictions near ground-truth centers. A parallel sub-pixel offset head compensates for accuracy lost through output stride discretization. We adopt this same dual-head architecture.

	Our feature extraction builds on self-supervised vision transformers. DINOv2 \cite{dinov2} demonstrated that frozen ViT features trained with a self-supervised objective transfer effectively to dense prediction tasks using only lightweight heads, a paradigm well suited to small datasets. We use DINOv3 \cite{dinov3}, the latest model in this family, with improved performance.

	A ViT-L/16 on $512 \times 512$ input produces only a $32 \times 32$ feature grid, too coarse for our sub-millimeter localization. AnyUp \cite{anyup} addresses this through guided feature upsampling: given low-resolution features and a high-resolution guide image, it produces high-resolution feature maps that preserve semantic content while achieving spatial precision from the guide. AnyUp is feature-agnostic and encoder-agnostic, generalizing without task-specific fine-tuning.

	\section{Methods}

	\paragraph{Dataset}
	Our dataset consists of 48 high-resolution photographs of 40\,cm indoor archery target faces, collected during practice sessions. Each image captures a target after shooting, with anywhere from 3 to over 100 arrow punctures visible. Images are taken at varying angles, producing perspective distortion. We manually annotated each image with point labels for every arrow shaft center in the rectified coordinate system, yielding 5,084 labeled punctures across the dataset. We partition the data using three-fold cross-validation with a fixed random seed for reproducibility.

	Given the limited dataset size, we employ aggressive data augmentation applied jointly to images and keypoint annotations via the Albumentations library \cite{albumentations}. Geometric augmentations include full $360^{\circ}$ rotation (exploiting the rotational symmetry of the target face), affine transforms with $\pm 10\%$ scaling, $\pm 5\%$ translation, and $\pm 5^{\circ}$ shear, as well as horizontal and vertical flips. Photometric augmentations include color jitter ($\pm 20\%$ brightness, contrast, and saturation; $\pm 10\%$ hue), random gamma adjustment, Gaussian blur (kernel 3--7\,px), motion blur, defocus blur, additive Gaussian noise ($\sigma = 0.012$--$0.028$), and random lighting effects (shadows and sun flare). These augmentations simulate the variability of real-world photographs in lighting conditions, camera shake, and image quality, and are critical for preventing overfitting when training on only a few dozen images.

	\paragraph{Canonical Rectification Pipeline}
	A central contribution of our pipeline is a canonical rectification stage that transforms each perspective-distorted photograph into a standardized coordinate system before the neural network processes it. This eliminates geometric variability that would otherwise need to be learned, simplifying the detection task and enabling annotations in a physical coordinate system where pixel distances correspond to known measurements.

	The rectification exploits the known color structure of the target face and operates in CIELAB color space, which provides robustness to illumination variation. We identify four diagnostic color boundaries: yellow-to-red (40\,mm radius), red-to-blue (80\,mm), blue-to-black (120\,mm), and black-to-white (160\,mm). For each boundary, binary region masks are created using adaptive thresholds on luminance and chroma channels with hue-angle constraints. Convex hulls are fitted to the region masks, and ellipses are fitted to the hull contours. A confidence-based scoring system selects the best nested set of boundary ellipses, enforcing the constraint that inner ellipses are contained within outer ones.

	This region-based approach is chosen over edge-based alternatives, which are fragile where arrow punctures disrupt ring boundaries.

\begin{figure}[t]
		\centering
		\includegraphics[width=0.48\columnwidth]{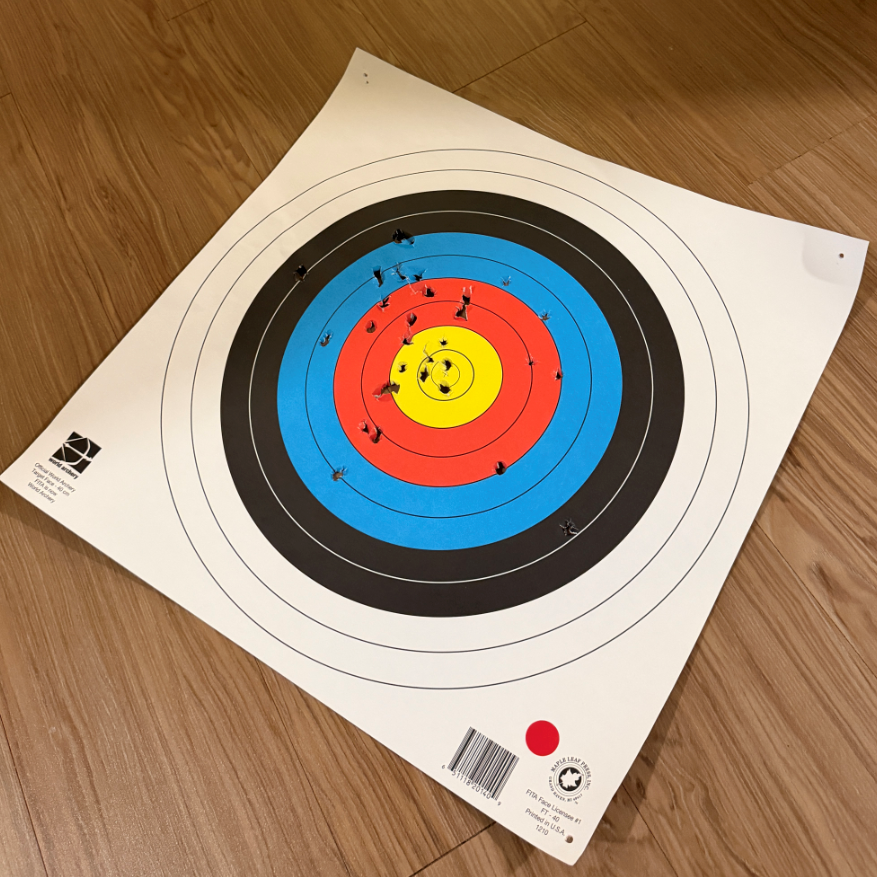}\hfill
		\includegraphics[width=0.48\columnwidth]{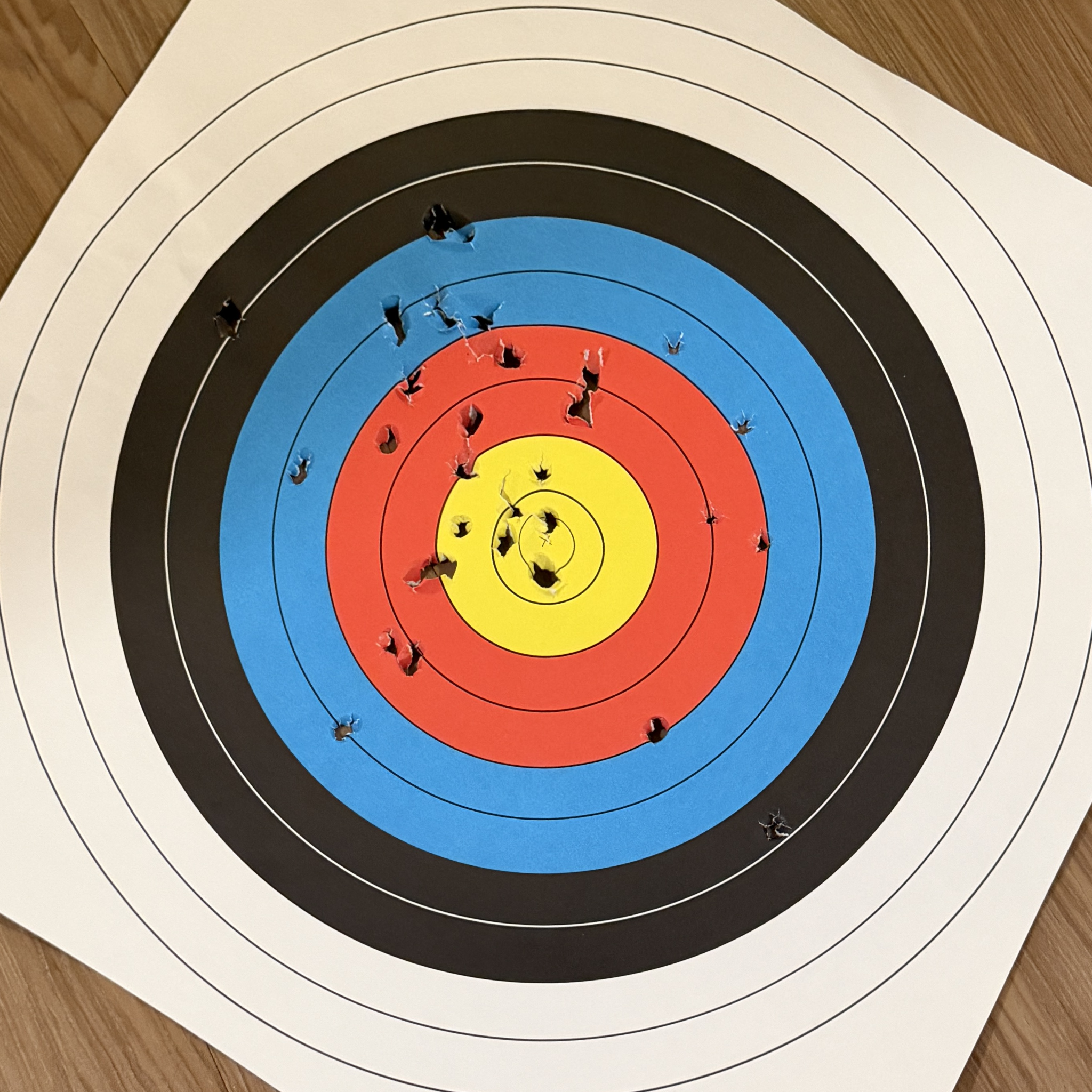}
		\caption{Input photograph (left) and canonically rectified output (right). The rectification pipeline maps the perspective-distorted target into a canonical coordinate system where scoring rings are concentric circles.}
		\label{fig:rectification}
	\end{figure}

	From the detected ellipses, per-angle radii are estimated for each ring and intermediate contours are interpolated between detected boundaries. A radial warp then transforms the image into a canonical $2048 \times 2048$ output where the target center is always at the image center, rings are concentric circles, and scale is uniform at approximately 0.195\,mm/pixel. Fig.~\ref{fig:rectification} shows the result of this process.

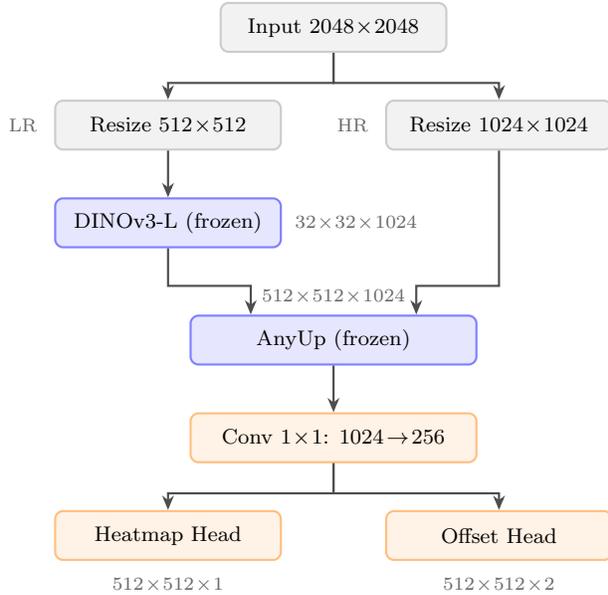
\begin{figure}[t]
		\centering
		\begin{tikzpicture}[
			x=1cm, y=1cm,
			block/.style={
				rectangle, draw, rounded corners=3pt,
				minimum width=3.0cm, minimum height=0.65cm,
				text centered, font=\footnotesize, inner sep=3pt,
				thick
			},
			frozen/.style={block, fill=blue!10, draw=blue!50},
			trainable/.style={block, fill=orange!10, draw=orange!50},
			data/.style={block, fill=gray!10, draw=gray!40},
			arr/.style={-{Stealth[length=5.5pt, width=4.5pt]}, thick, draw=black!70},
			lbl/.style={font=\scriptsize, text=black!60},
		]
		\def\vsep{1.3}
		\def\hsep{2.2}

		\node[data] (input) at (0, 0) {Input $2048\!\times\!2048$};

		\node[data] (lr) at (-\hsep, -\vsep) {Resize $512\!\times\!512$};
		\node[data] (hr) at (\hsep, -\vsep) {Resize $1024\!\times\!1024$};

		\node[lbl, left=0.1cm of lr] {LR}; 
		\node[lbl, left=0.1cm of hr] {HR};

		\node[frozen] (dino) at (-\hsep, -2*\vsep) {DINOv3-L (frozen)};
		\node[lbl, right=0.05cm of dino] {$32\!\times\!32\!\times\!1024$};

		\node[frozen, minimum width=3.8cm] (anyup) at (0, -3.2*\vsep) {AnyUp (frozen)};
		\node[lbl, above=0.02cm of anyup] {$512\!\times\!512\!\times\!1024$};

		\node[trainable, minimum width=3.8cm] (proj) at (0, -4.2*\vsep) {Conv $1\!\times\!1$: $1024\!\to\!256$};

		\node[trainable] (hmap) at (-\hsep, -5.2*\vsep) {Heatmap Head};
		\node[trainable] (off) at (\hsep, -5.2*\vsep) {Offset Head};

		\node[lbl, below=0.1cm of hmap] {$512\!\times\!512\!\times\!1$};
		\node[lbl, below=0.1cm of off] {$512\!\times\!512\!\times\!2$};

		\draw[arr] (input.south) -- ++(0,-0.4) -| (lr.north);
		\draw[arr] (input.south) -- ++(0,-0.4) -| (hr.north);
		\draw[arr] (lr.south) -- (dino.north);
		\draw[arr] (dino.south) -- ++(0,-0.5) -| ([xshift=-1.1cm]anyup.north);
		\draw[arr] (hr.south) -- ++(0,-1.8) -| ([xshift=1.1cm]anyup.north);
		\draw[arr] (anyup.south) -- (proj.north);
		\draw[arr] (proj.south) -- ++(0,-0.4) -| (hmap.north);
		\draw[arr] (proj.south) -- ++(0,-0.4) -| (off.north);
		\end{tikzpicture}
		\caption{Model architecture. Blue blocks are frozen pretrained modules (304\,M params: 303\,M backbone + 1\,M upsampler); orange blocks contain the 3.8\,M trainable parameters.}
		\label{fig:architecture} 
	\end{figure}

	\paragraph{Model Architecture}
	The detection model has three stages: frozen feature extraction, guided upsampling, and task-specific prediction heads. Fig.~\ref{fig:architecture} illustrates the complete pipeline. Of the model's 308\,M total parameters, only 3.8\,M are trainable (the projection layer and detection heads), keeping capacity appropriate for our small dataset.

	\textit{Feature extraction.} A DINOv3 ViT-L/16 \cite{dinov3}, pretrained with self-supervised learning, serves as the backbone. The $2048 \times 2048$ rectified image is resized to $512 \times 512$, producing a $32 \times 32$ grid of 1024-dimensional patch tokens (after dropping the CLS and register tokens). The backbone is entirely frozen during training.

	\textit{Guided upsampling.} AnyUp \cite{anyup} takes the $32 \times 32 \times 1024$ features and a $1024 \times 1024$ version of the rectified image as a high-resolution guide, producing $512 \times 512 \times 1024$ upsampled features. We use a $1024 \times 1024$ guide rather than the full $2048 \times 2048$ image because AnyUp's internal image encoder and RoPE modules run convolutions at full guide resolution, causing out-of-memory errors at $2048 \times 2048$ on consumer GPUs. AnyUp is frozen. A learnable $1 \times 1$ convolution projects the 1024-channel features to 256 channels.

	\textit{Detection heads.} Two parallel convolutional heads operate on the $512 \times 512 \times 256$ features. The \textbf{heatmap head} consists of three Conv--BN--ReLU blocks (each $3 \times 3$, 256 channels) followed by a $1 \times 1$ convolution producing a single-channel map of arrow-center logits. Its final bias is initialized to $-2.19$ so that $\sigma(-2.19) \approx 0.1$, preventing initial predictions from overwhelming the focal loss. The \textbf{offset head} shares the same three-block architecture but outputs two channels $(dx, dy)$ predicting sub-pixel offsets to the nearest arrow center, with weights zero-initialized.

	\paragraph{Loss Functions}
	The heatmap head is trained with a Gaussian focal loss adapted from CenterNet \cite{centernet}. Ground-truth heatmaps are generated by rendering a Gaussian blob ($\sigma = 6.6$\,px, radius $= 4\sigma$) at each annotated arrow center; where blobs overlap, the element-wise maximum is taken. Let $\hat{Y} = \sigma(Z)$ denote the predicted probability and $Y$ the ground-truth value at a pixel. The loss is:
	\begin{align}
		L_{\text{hm}} &= \frac{-1}{N} \sum_{xy}
		\begin{cases}
			(1\!-\!\hat{Y})^{\alpha} \log \hat{Y}
			& \text{if } Y \!\ge\! 0.99 \\[3pt]
			\hat{Y}^{\alpha} (1\!-\!Y)^{\beta} \log(1\!-\!\hat{Y})
			& \text{else}
		\end{cases}
		\label{eq:focal}
	\end{align}
	where $N$ is the number of positive pixels, $\alpha = 3.86$ controls the focal weight on hard examples, and $\beta = 2.92$ downweights easy negatives near but not at ground-truth centers. Both $\alpha$ and $\beta$ were selected by hyperparameter optimization.

	The offset head is supervised only within a radius of 24\,px around each arrow center (the product of the arrow shaft radius and an offset radius factor of 2.0). Outside this mask, offset predictions incur no loss. Within the mask, a smooth $L_1$ loss with transition threshold $\beta_{\text{off}} = 1.0$ is applied:
	\begin{equation}
		L_{\text{off}} = \frac{1}{|\mathcal{M}|} \!\sum_{(x,y) \in \mathcal{M}} \!\text{SmL1}(\hat{O}_{xy} \!-\! O_{xy},\; \beta_{\text{off}})
	\end{equation}
	where $\mathcal{M}$ is the set of masked pixels, $\hat{O}_{xy}$ is the predicted offset, and $O_{xy}$ is the ground-truth offset pointing from pixel $(x,y)$ to the nearest arrow center. The total loss is $L = L_{\text{hm}} + 0.1 \cdot L_{\text{off}}$.

	\paragraph{Training Procedure}
	We optimize the 3.8\,M trainable parameters with AdamW \cite{adamw} using a learning rate of $1.4 \times 10^{-4}$ and weight decay $2.8 \times 10^{-5}$. The learning rate follows a cosine annealing schedule over 100 epochs, decaying to a minimum of $1.4 \times 10^{-6}$. Gradients are clipped to a maximum norm of 1.0. Training uses automatic mixed precision (AMP) with a batch size of 1 due to the limited dataset. The entire model fits comfortably on a consumer GPU (peak VRAM 6.8\,GB on an RTX 3080 12\,GB); the final 100-epoch training run completed in approximately 30 minutes on an NVIDIA H200.

	\paragraph{Hyperparameter Optimization}
	We employ a bi-level optimization strategy to jointly tune training hyperparameters and post-hoc decoder settings. The outer level uses Optuna \cite{optuna} with a Tree-structured Parzen Estimator (TPE) sampler to search over six training hyperparameters: learning rate (log-uniform $[10^{-5}, 10^{-3}]$), weight decay (log-uniform $[10^{-6}, 10^{-3}]$), focal loss exponents $\alpha \in [1, 4]$ and $\beta \in [2, 6]$, Gaussian blob sigma $\sigma \in [2, 10]$\,px, and an augmentation intensity flag. Each trial trains the model for up to 30 epochs with early stopping (patience of 15 epochs on validation loss). A MedianPruner with 5 startup trials terminates unpromising trials early: of 39 total trials, 22 were pruned before completion. Each completed trial evaluates all three cross-validation folds; the objective is the mean F1 score across folds.

	The inner level performs an exhaustive grid search over decoder parameters (detection threshold over 10 values from 0.01 to 0.4, and NMS kernel size over 15 odd values from 3 to 31) after each trial's training completes. This decouples the decoder from the training dynamics, ensuring that the Optuna objective reflects the best achievable F1 for each set of learned weights.

	The sweep ran for 12.5 hours on an NVIDIA H200. Table~\ref{tab:hpo_space} summarizes the search space and the selected values from the best trial (mean F1 = 0.877 across folds at 30 epochs). The decoder configuration from the best trial (threshold $= 0.5$, NMS kernel $= 15$) is fixed for all subsequent evaluation. Three final models were then trained from scratch, one per fold, with the selected hyperparameters for 100 epochs each.

	\begin{table}[t]
		\centering
		\caption{Hyperparameter search space and selected values.}
		\label{tab:hpo_space}
		\small
		\begin{tabular}{@{}llr@{}}
			\toprule
			\textbf{Parameter} & \textbf{Search Range} & \textbf{Selected} \\
			\midrule
			Learning rate & $[10^{-5},\, 10^{-3}]$ (log) & $1.4 \!\times\! 10^{-4}$ \\
			Weight decay & $[10^{-6},\, 10^{-3}]$ (log) & $2.8 \!\times\! 10^{-5}$ \\
			Focal $\alpha$ & $[1.0,\, 4.0]$ & 3.86 \\
			Focal $\beta$ & $[2.0,\, 6.0]$ & 2.92 \\
			Heatmap $\sigma$ & $[2.0,\, 10.0]$\,px & 6.6 \\
			Strong augmentation & \{true, false\} & false \\
			\bottomrule
		\end{tabular}
	\end{table}

	\paragraph{Evaluation Protocol}
	At inference, the heatmap logits are passed through a sigmoid, followed by $k \times k$ max-pooling non-maximum suppression (NMS) to extract local peaks above a confidence threshold. Detected peaks are optionally refined by the offset head's predicted sub-pixel displacements. A minimum-separation filter removes duplicate detections closer than a specified radius.

	For evaluation, predicted and ground-truth arrow centers are matched using the Hungarian algorithm \cite{kuhn1955} on the pairwise Euclidean distance matrix. A prediction--ground-truth pair is counted as a true positive (TP) if their distance is at most 15\,px (11.7\,mm in the $512 \times 512$ output space). Precision, recall, and F1 score are computed from the matched counts. Localization accuracy is measured as the mean Euclidean distance of matched TP pairs. At $512 \times 512$ output resolution, each pixel corresponds to $400 / 512 \approx 0.78$\,mm on the target face.

	For scoring, each detected arrow center's distance from the target center is converted to millimeters. The line-cutting rule is applied by subtracting the arrow shaft radius (2.25\,mm for 4.5\,mm micro-diameter shafts) from this distance, then awarding the highest ring score whose outer boundary exceeds the effective distance.

	\section{Results}

\begin{figure}[t]
		\centering
		\includegraphics[width=\columnwidth]{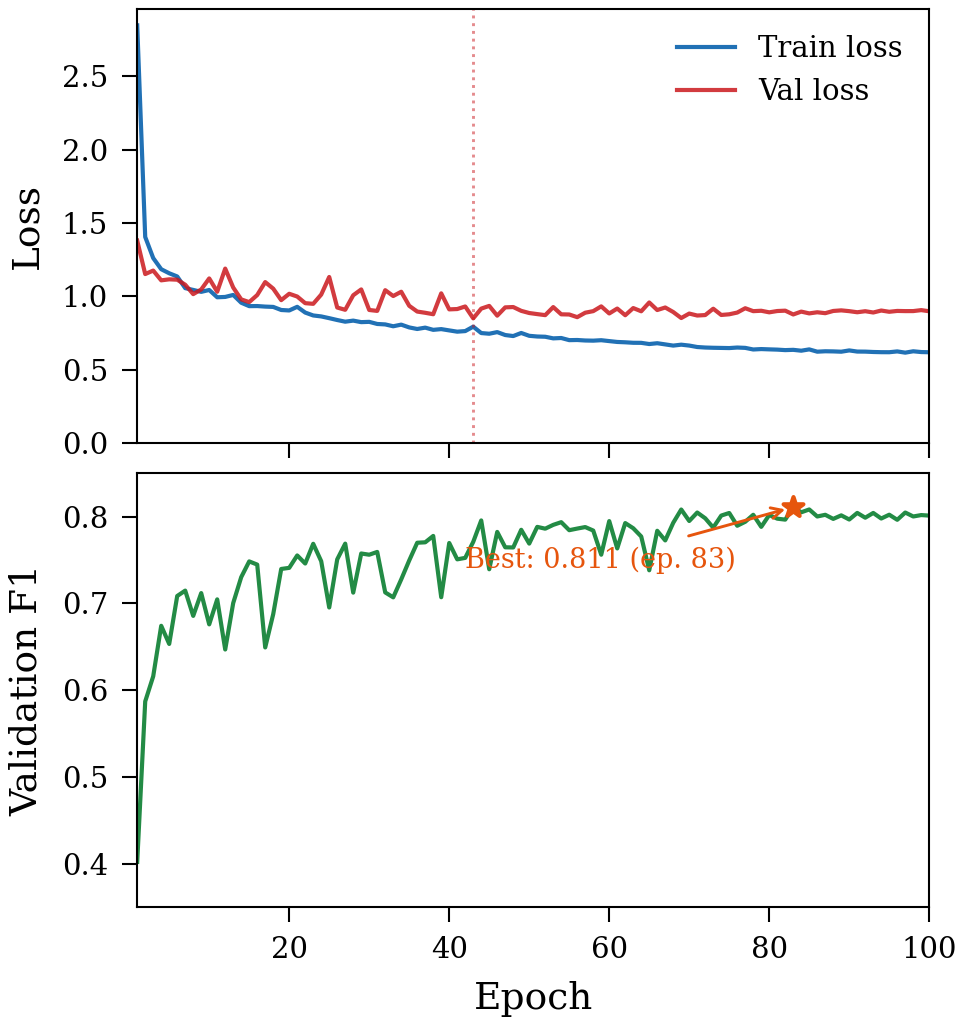}
		\caption{Training and validation loss (top) and validation F1 score (bottom) over 100 epochs for fold~1. The star marks the best F1 (0.811 at epoch 83). The dotted line marks the epoch of lowest validation loss.}
		\label{fig:training_curves}
	\end{figure}

	\paragraph{Training Convergence}
	Fig.~\ref{fig:training_curves} shows representative training and validation loss curves alongside the validation F1 score over 100 epochs for one fold. Training loss decreases monotonically from 2.85 to 0.62, while validation loss decreases from 1.38 to 0.90 with a best of 0.85 at epoch 43. The validation F1 score rises from 0.40 to a plateau around 0.80, reaching its peak of 0.811 at epoch 83. Over the final 10 epochs, F1 varies by only $\pm 0.003$, indicating stable convergence. The moderate gap between training and validation loss suggests mild overfitting, but the continued rise and plateau of the F1 score confirms that the model generalizes. All three folds exhibit similar convergence behavior.

	\paragraph{Decoder Settings}
	The decoder parameters (detection threshold, NMS kernel size, minimum arrow separation, and offset refinement) were selected during hyperparameter optimization on 30-epoch models evaluated across all three folds. To avoid optimistic bias from re-tuning these parameters on the final 100-epoch validation data, we fix the decoder configuration at the values selected during HPO (threshold $= 0.5$, NMS kernel $= 15$, separation radius $= 4$\,px) and apply it uniformly across all folds when reporting final metrics.

	\paragraph{Three-Fold Cross-Validation Results}
	Table~\ref{tab:cv_results} reports detection and localization metrics across all three folds of the final 100-epoch model using the fixed decoder configuration.

	\begin{table}[t]
		\centering
		\caption{Final model performance across three cross-validation folds with fixed decoder settings (threshold $= 0.5$, NMS $= 15$, no offsets).}
		\label{tab:cv_results}
		\footnotesize
		\setlength{\tabcolsep}{4pt}
		\begin{tabular}{@{}lcccc@{}}
			\toprule
			 & F1 & Prec. & Recall & Err.\,(mm) \\
			\midrule
			Fold 0 & .906 & .948 & .867 & 1.33 \\
			Fold 1 & .879 & .941 & .825 & 1.42 \\
			Fold 2 & .893 & .963 & .832 & 1.48 \\
			\midrule
			Mean\,$\pm$\,Std & $.893\!\pm\!.011$ & $.951\!\pm\!.009$ & $.841\!\pm\!.018$ & $1.41\!\pm\!.06$ \\
			\bottomrule
		\end{tabular}
	\end{table}

\begin{figure}[t]
		\centering
		\includegraphics[width=\columnwidth]{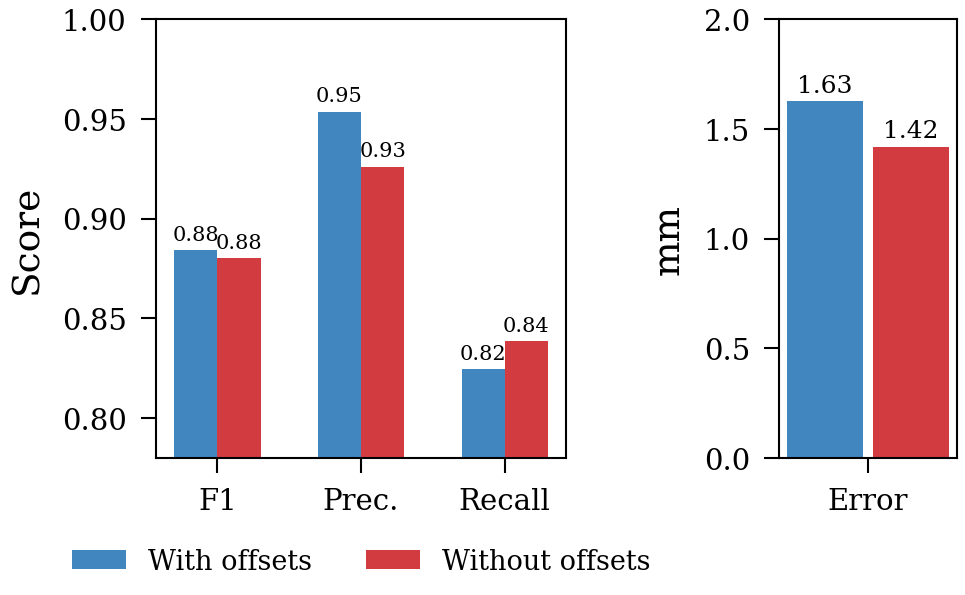}
		\caption{Ablation study comparing detection metrics (left) and mean localization error (right) with and without offset refinement.}
		\label{fig:ablation}
	\end{figure}

	\paragraph{Ablation: Offset Head}
	To evaluate the contribution of the offset regression head, we compare the decoder with offsets enabled against offsets disabled, averaged across all three folds (Fig.~\ref{fig:ablation}). Table~\ref{tab:ablation} summarizes the results.

	\begin{table}[t]
		\centering
		\caption{Ablation study: offset head contribution (mean $\pm$ std across 3 folds).}
		\label{tab:ablation}
		\footnotesize
		\setlength{\tabcolsep}{4pt}
		\begin{tabular}{@{}lcccc@{}}
			\toprule
			Offsets & F1 & Prec. & Recall & Err.\,(mm) \\
			\midrule
			Yes     & $.893\!\pm\!.009$ & $.952\!\pm\!.009$ & $.842\!\pm\!.018$ & $1.56\!\pm\!.06$ \\
			No      & $.893\!\pm\!.011$ & $.951\!\pm\!.009$ & $.841\!\pm\!.018$ & $1.41\!\pm\!.06$ \\
			\midrule
			$\Delta$ & $<$.001 & $+$.001 & $+$.001 & $+$.15 \\
			\bottomrule
		\end{tabular}
	\end{table}

	Across all three folds, the offset head provides negligible F1 improvement ($< 0.001$) while increasing mean localization error by 0.15\,mm ($1.56 \pm 0.06$ vs.\ $1.41 \pm 0.06$\,mm). The F1 difference is an order of magnitude smaller than the cross-fold standard deviation ($\pm 0.011$), indicating that the offset head's contribution is not meaningful. This suggests that the heatmap peaks are already well-localized at the training resolution and offset regression on a small dataset introduces noise rather than useful refinement. We hypothesize that the Gaussian target blobs with $\sigma = 6.6$\,px provide sufficient gradient signal to localize peaks accurately. Given the clear localization advantage, we recommend the heatmap-only configuration for deployment.

	\paragraph{Comparison with Prior Work}
	The most directly comparable system is Kim et al.\ \cite{kim2025}, which reports a mean localization error of 1.77\,mm on outdoor archery targets using a fully supervised convolutional network. Our heatmap-only configuration achieves a mean localization error of $1.41 \pm 0.06$\,mm across three folds. The two results are not directly comparable: the target types differ (40\,cm indoor vs.\ 122\,cm outdoor), the images come from different capture conditions, and the evaluation protocols are not identical. Within these caveats, the results suggest that transfer learning from frozen self-supervised features can achieve competitive localization accuracy with far fewer training images and no task-specific backbone training.

\begin{figure*}[!htb]
		\centering
		\includegraphics[width=0.48\textwidth]{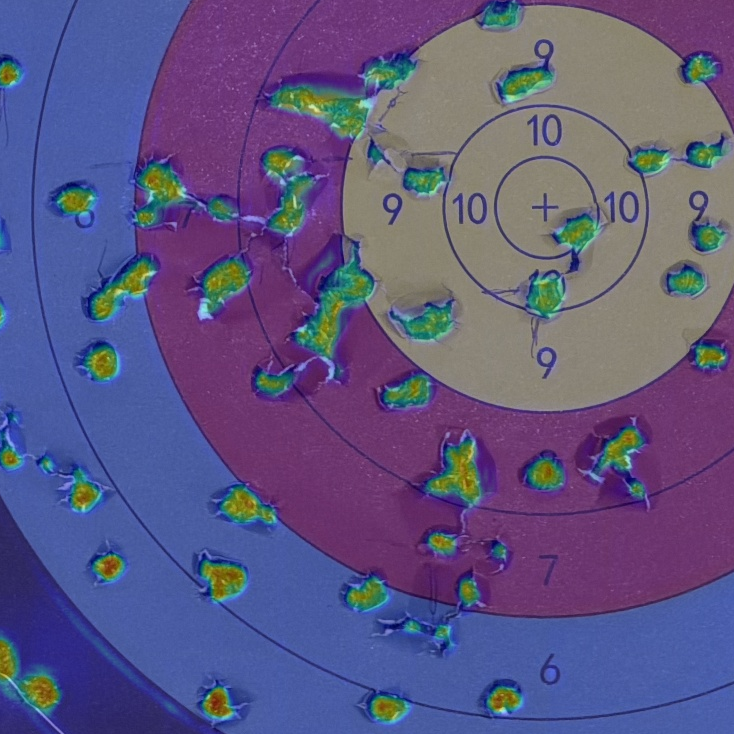}\hfill
		\includegraphics[width=0.48\textwidth]{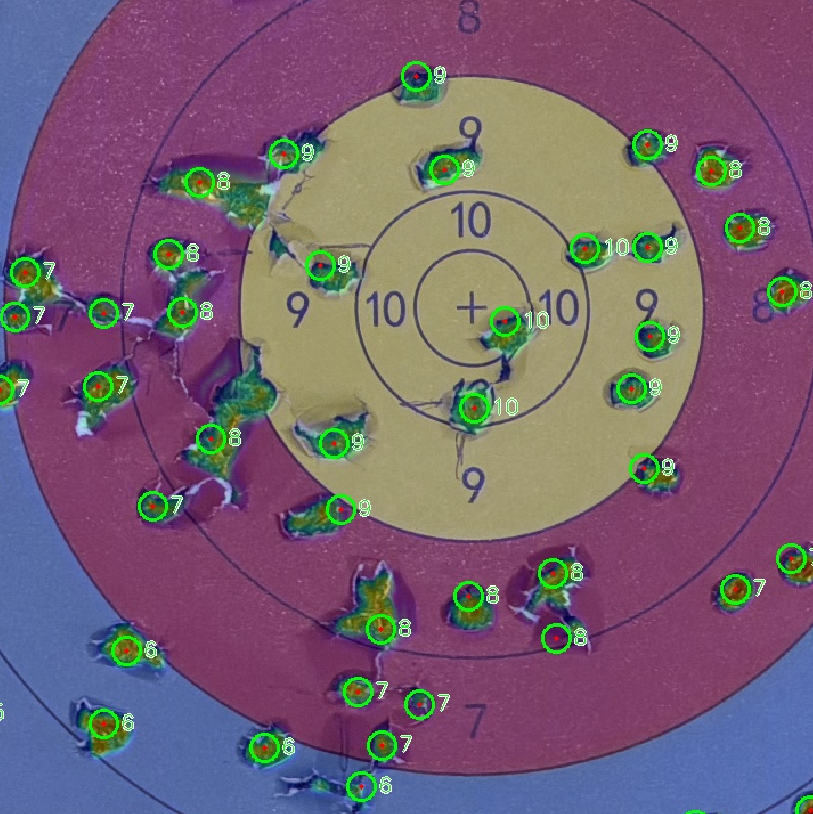}
		\caption{Qualitative results on a validation image. Left: predicted heatmap overlay showing confident peaks at arrow locations. Right: final detections after NMS with per-arrow ring scores.}
		\label{fig:qualitative}
	\end{figure*}

	\paragraph{Qualitative Results}
	Fig.~\ref{fig:qualitative} shows the model's predicted heatmap overlaid on a validation image alongside the resulting detections after NMS decoding. The heatmap produces sharp, well-localized peaks at arrow locations across all scoring rings. The model successfully detects arrows in cluttered regions where multiple punctures are tightly grouped, though a small number of closely overlapping impacts remain challenging.

\begin{figure}[!htb]
		\centering
		\includegraphics[width=0.85\columnwidth]{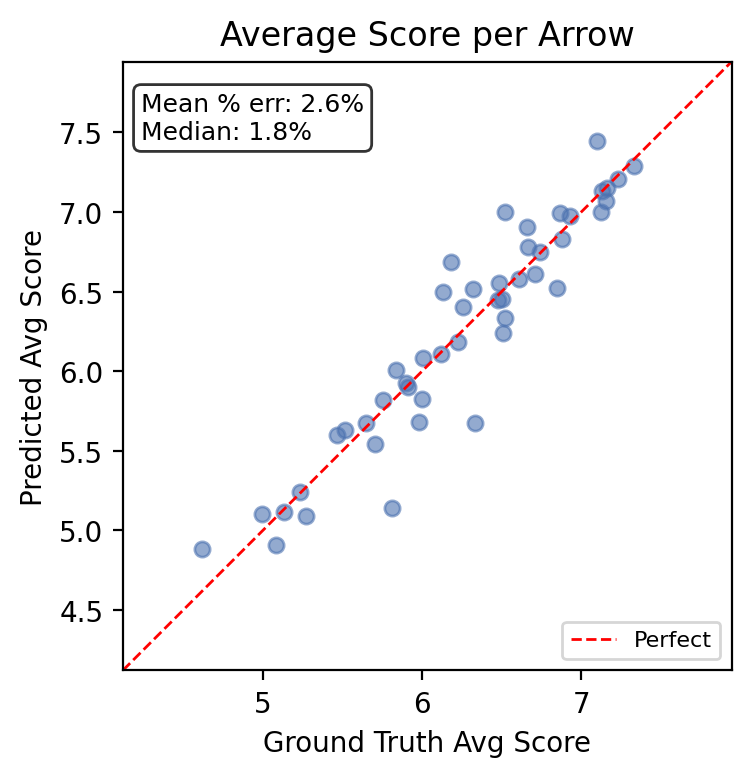}
		\caption{Predicted vs.\ ground-truth average arrow score across all validation images. Points near the diagonal indicate accurate score recovery.}
		\label{fig:archery_metrics}
	\end{figure}

	\paragraph{Downstream Archery Metrics}
	To evaluate the system's utility for its intended application, we compute two archery-specific metrics across all validation images (Fig.~\ref{fig:archery_metrics}). First, we compare the \emph{average arrow score} derived from the model's predicted positions against the ground-truth average score for each image. The median per-image error is 1.8\% (mean 2.6\%), with a mean signed error of $-0.001$ points, indicating that the system recovers the average score with negligible bias. Second, we measure the \emph{group centroid error}: the Euclidean distance between the centroid of all predicted arrow positions and the centroid of all ground-truth positions, converted to millimeters. The median centroid error is 4.00\,mm (mean $4.90 \pm 3.65$\,mm), well within a single ring width (20\,mm) and sufficient for identifying systematic aiming bias.

	\section{Discussion}

	\paragraph{Effectiveness of Frozen Backbones for Small Datasets}
	A central finding of this work is that a frozen self-supervised vision transformer, combined with guided feature upsampling and lightweight trainable heads, achieves strong detection and localization performance on a dataset of only 48 images. The DINOv3-Large backbone provides 303\,M parameters' worth of visual representation capacity without any risk of overfitting on our small training set. The 3.8\,M trainable parameters are sufficient to adapt these general-purpose features to the specific task of arrow puncture detection. This validates the paradigm of using frozen foundation models as feature extractors for specialized, low-data vision tasks.

	\paragraph{Value of Geometric Preprocessing}
	The canonical rectification pipeline is essential to our approach. By transforming all images into a standardized coordinate system \emph{before} any learning occurs, we eliminate perspective distortion as a source of variability. Without rectification, the model would need to implicitly learn perspective invariance from only 48 images, which is impractical. The rectification also enables direct physical interpretation of detection coordinates: pixel distances in the rectified frame correspond to known millimeter measurements, allowing automatic scoring without additional calibration. The affine shear augmentation provides mild geometric perturbation so that the model remains robust to residual imperfections in the rectification output.

	\paragraph{The Offset Head Paradox}
	In conventional CenterNet-style detectors operating on natural images, the offset head provides meaningful sub-pixel refinement because the output stride of the feature extractor introduces systematic discretization error. In our architecture, AnyUp's guided upsampling already recovers much of the spatial precision lost by the backbone's patch tokenization. With a Gaussian target of $\sigma = 6.6$\,px, the focal loss drives heatmap peaks to align closely with true centers. The offset head, trained with only a masked $L_1$ loss on a small dataset, appears to learn a noisier correction signal than the heatmap alone provides. This suggests that offset regression may be less beneficial when the feature upsampling already resolves the output to fine spatial granularity, particularly in low-data regimes where offset predictions are harder to learn reliably.

	\paragraph{Failure Modes}
	Error analysis reveals two dominant failure modes. First, \emph{compound punctures}, where multiple arrows strike within one shaft diameter of each other, produce overlapping tears whose Gaussian heatmap responses merge under NMS, causing the detector to emit a single peak where two or more arrows are present. This is the primary source of missed detections and accounts for the gap between precision and recall. Second, heavily worn target faces with accumulated paper damage can produce false positives: large tears or peeled paper regions occasionally resemble puncture signatures in the feature space. These cases are rare but illustrative of what an F1 of $\sim$0.89 looks like in practice: the detector is reliable on isolated punctures but struggles with the combinatorial ambiguity of overlapping damage.

	\paragraph{Limitations}
	Several limitations should be noted. The dataset of 48 images comes from a single archer and venue, which may limit generalization to different lighting conditions, camera equipment, or target wear patterns. The system is validated only on the standard 40\,cm indoor target face; different target sizes or outdoor target faces would require modifications to the rectification pipeline's color boundary definitions. Although we report three-fold cross-validation results, the dataset does not include a held-out test set that is entirely unseen during model selection. The decoder hyperparameters were selected during HPO on 30-epoch models and fixed for final evaluation, which mitigates but does not eliminate the risk of optimistic bias from using the same folds for both selection and reporting. A dedicated test partition (even 5--6 images) would provide stronger evidence of generalization. Finally, the NMS-based evaluation protocol with a fixed 15-pixel matching radius may not capture all nuances of localization quality.

	\section{Conclusion}

	We presented a system for automatic detection, localization, and scoring of arrow punctures from photographs of 40\,cm indoor archery targets. By combining a color-based canonical rectification pipeline with frozen DINOv3 features, AnyUp guided upsampling, and lightweight CenterNet-style detection heads, we achieve a mean F1 score of $0.893 \pm 0.011$ and a mean localization error of $1.41 \pm 0.06$\,mm across three cross-validation folds with the recommended heatmap-only decoder, using only 3.8\,M trainable parameters on a 48-image dataset. An ablation study reveals that the offset regression head provides marginal detection improvement at the cost of localization accuracy, suggesting that guided upsampling already resolves the spatial precision lost by the backbone's patch tokenization. These results demonstrate that frozen self-supervised foundation models with minimal task-specific adaptation offer a practical paradigm for dense prediction in small-data regimes, though validation on a held-out test set and broader target conditions remains future work.


\begin{thebibliography}{99}
	\bibitem{centernet}
	X.~Zhou, D.~Wang, and P.~Kr\"ahenb\"uhl, ``Objects as points,'' \textit{arXiv preprint arXiv:1904.07850}, 2019.

	\bibitem{dinov2}
	M.~Oquab \textit{et al.}, ``DINOv2: Learning robust visual features without supervision,'' \textit{Trans. Mach. Learn. Res.}, 2024.

	\bibitem{dinov3}
	O.~Sim\'eoni \textit{et al.}, ``DINOv3,'' \textit{arXiv preprint arXiv:2508.10104}, 2025.

	\bibitem{anyup}
	T.~Wimmer, P.~Truong, M.-J.~Rakotosaona, M.~Oechsle, F.~Tombari, B.~Schiele, and J.~E.~Lenssen, ``AnyUp: Universal feature upsampling,'' in \textit{Proc. Int. Conf. Learn. Represent. (ICLR)}, 2026.

	\bibitem{kim2025}
	S.~Kim, J.~Moon, and E.~C.~Lee, ``Archery score analysis system for outdoor environments,'' \textit{Proc. Inst. Mech. Eng., Part P: J. Sports Eng. Technol.}, 2025.

	\bibitem{worldarchery}
	World Archery, ``Rulebook, Book 3: Target Archery,'' World Archery Federation, 2025. [Online]. Available: \url{https://www.worldarchery.sport/rulebook/article/13}

	\bibitem{optuna}
	T.~Akiba, S.~Sano, T.~Yanase, T.~Ohta, and M.~Koyama, ``Optuna: A next-generation hyperparameter optimization framework,'' in \textit{Proc. ACM SIGKDD Int. Conf. Knowl. Disc. Data Mining}, 2019, pp.~2623--2631.

	\bibitem{albumentations}
	A.~Buslaev, V.~I.~Iglovikov, E.~Khvedchenya, A.~Parinov, M.~Druzhinin, and A.~A.~Kalinin, ``Albumentations: Fast and flexible image augmentations,'' \textit{Information}, vol.~11, no.~2, p.~125, 2020.

	\bibitem{adamw}
	I.~Loshchilov and F.~Hutter, ``Decoupled weight decay regularization,'' in \textit{Proc. Int. Conf. Learn. Represent. (ICLR)}, 2019.

	\bibitem{kuhn1955}
	H.~W.~Kuhn, ``The Hungarian method for the assignment problem,'' \textit{Naval Research Logistics Quarterly}, vol.~2, no.~1--2, pp.~83--97, 1955.
	\end{thebibliography}
\end{document}